\documentclass[conference]{IEEEtran}
\usepackage{cite}
\usepackage{amsmath,amssymb}
\usepackage{booktabs}
\usepackage{xcolor}
\usepackage{url}
\usepackage{tikz}
\usetikzlibrary{positioning}

\newcommand{\code}[1]{\texttt{\small #1}}

\begin{document}

\title{Agent Recovery Gone Wrong: Measuring Harm Beyond Duplicate Calls}

\author{\IEEEauthorblockN{Artem Trofimov}
\IEEEauthorblockA{AVIV Group\\Berlin, Germany\\artemtrofimov@acm.org}
\and
\IEEEauthorblockN{Boris Novikov}
\IEEEauthorblockA{Independent Researcher\\Helsinki, Finland\\borisnov@acm.org}}

\maketitle
\bstctlcite{IEEEexample:BSTcontrol}

\begin{abstract}
Code-executing LLM agents submit blocks of code that issue several API writes. If a block fails after some of its writes have taken effect, the agent rewrites the block and continues. It is unknown how often such a recovery harms the external state when the failure occurs naturally in the code written by the agent. In this paper, we measure it on recorded trajectories of AppWorld agents. In a stratified random sample of 200 candidate episodes across ten models, 145 are partial executions confirmed by the database state. In the sample, 30 of 143 confirmed partial executions with a determinable outcome are followed by a harmful recovery, and its violation remains at the end of the run, which gives a stratification-weighted estimate of 24.6\%. Two thirds of the harmful recoveries duplicate an effect, and the rest delete user data or make a wrong first-time write, e.g., an order of a wrong product. A rule that searches for repeated calls finds 75\% of duplications but only 41\% of all harm. We also examine two sources of operation identity for an idempotency contract. In a small pilot (17 episodes, one local model), identifiers supplied by the model do not detectably reduce harm, because the model issues a new identifier for the same operation. Identity derived from the call content, evaluated counterfactually on the sample, would have matched every harmful write in 14 to 18 of the 30 harmful recoveries, but in none of those with a wrong first-time write, and it would also have matched writes required by the task in 30 episodes. Based on these results, we outline design requirements for safer agent recovery.
\end{abstract}


\section{Introduction}

Agents that write and execute code change external services without a transaction. A single block can add an item to a cart, create an address, and place an order. If the block fails on the third call, the first two calls have already taken effect. The agent receives only an error, rewrites the block, and runs it again. We call this scenario a \emph{recovery after partial execution}. It does not require a crash or a lost response: an ordinary exception in generated code is enough, so any deployed agent with write access to APIs can face it.

Recent systems consider the agent-tool boundary as a transactional one: they stage effects~\cite{chen2026cordon}, use idempotency keys and compensation~\cite{mohammadi2026atomix,perera2026rac}, verify ambiguous outcomes~\cite{mansoor2026verified}, or inject crashes around a tool call~\cite{mohammadi2026rubicon}. However, these results are obtained on injected faults. It remains unclear how often a recovery harms the external state when the failure occurs naturally in the code written by the agent.

Measuring this harm requires recorded executions and the state of the service before and after the recovery. AppWorld provides both: its agents write Python code against applications backed by databases, and its runs can be replayed deterministically~\cite{trivedi2024appworld}. In 13{,}988 AppWorld runs, static analysis identifies 1{,}897 failed blocks with at least one write call before the failing line. In 70.6\% of these cases, the agent rewrites the code and issues these writes again. We use these runs as a measurement instrument.

In this paper, we consider the following questions:
\begin{itemize}
\item[\textbf{Q1}] How often does a recovery after partial execution add harm to the external state?
\item[\textbf{Q2}] How much of this harm can repeat-based signals detect?
\item[\textbf{Q3}] Does an idempotency contract help, and where can it take the identity of an operation?
\end{itemize}

In summary, our contributions are as follows:
\begin{enumerate}
\item We estimate how often a recovery after a state-confirmed partial execution leaves a task violation and which kinds of harm it produces.
\item We compare repeat-based signals with labels based on state and task, and report the defects of our own instrument revealed by this comparison.
\item We examine two sources of operation identity for an idempotency contract: the model, in a pilot, and the call content, in a counterfactual analysis of the sample. The pilot is small, so we consider its results as emerging evidence.
\end{enumerate}

In the catalog of agent-tool anomalies~\cite{trofimov2026toolcallssucceedworkflows}, duplication follows the duplicated-effect pattern, although here the completed writes are observable in the state; the other kinds of harm are task-level errors outside that catalog. Data, labels, and scripts are available online.\footnote{\url{https://github.com/flame-stream/agent-recovery-harm}}

\section{Data and Labels}\label{sec:setting}

\textbf{Corpus and sample.} From the official and leaderboard experiment outputs of AppWorld, we build a registry of 1{,}917 candidate episodes. It contains 20 more episodes than the 1{,}897 blocks mentioned above, because the detector now recognizes the API wrapper of one more agent. A candidate episode is a block that raised an error with at least one write before the failing line. Older experiments dominate the registry. Therefore, we stratify it by the release year of the model before sampling (before 2025: 1{,}403 episodes, 27 configurations; 2025--2026: 514 episodes, 8 configurations) and draw a simple random sample of 100 episodes from each stratum with a fixed seed. The sample covers ten models. We sample episodes, not models, so its composition follows the registry, where a few configurations provide most of the failed blocks.

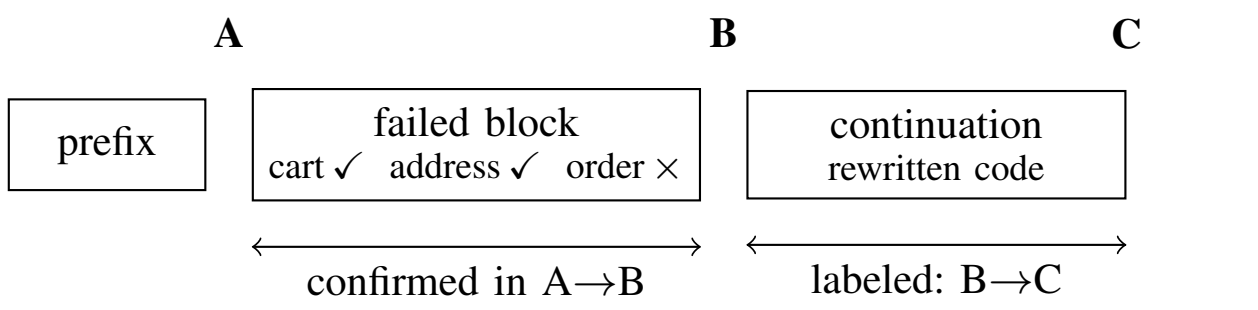
\begin{figure}[t]
\centering
\begin{tikzpicture}[font=\footnotesize, node distance=3mm,
  box/.style={draw, minimum height=6mm, inner xsep=3pt},
  snap/.style={font=\footnotesize\bfseries}]
\node[box, minimum width=13mm] (pre) {prefix};
\node[box, right=of pre, minimum width=29mm, align=center] (blk) {failed block\\[-1pt]{\scriptsize cart\,$\checkmark$\; address\,$\checkmark$\; order\,$\times$}};
\node[box, right=of blk, minimum width=25mm, align=center] (cont) {continuation\\[-1pt]{\scriptsize rewritten code}};
\node[snap, above=1.5mm of blk.north west, xshift=-1.5mm] (A) {A};
\node[snap, above=1.5mm of blk.north east, xshift=1.5mm] (B) {B};
\node[snap, above=1.5mm of cont.north east] (C) {C};
\draw[<->] ([yshift=-3mm]blk.south west) -- node[below]{confirmed in A$\to$B} ([yshift=-3mm]blk.south east);
\draw[<->] ([yshift=-3mm]cont.south west) -- node[below]{labeled: B$\to$C} ([yshift=-3mm]cont.south east);
\end{tikzpicture}
\caption{An episode and its snapshots. Partial execution is confirmed by A$\to$B, and harm is labeled by B$\to$C against the task.}
\label{fig:episode}
\end{figure}

\textbf{Replay and snapshots.} We re-execute each sampled episode from its recorded code with a hook on the API boundary. All 200 episodes are reproduced. As shown in Fig.~\ref{fig:episode}, we store database snapshots before the failed block (A), after it (B), and at the end of the trajectory (C). Further, we denote all actions of the original agent between B and C as a \emph{continuation}.

\textbf{Confirmed partial execution.} A candidate is confirmed if the difference between A and B contains a change that corresponds to a write counted by the detector as completed.

\textbf{Labels.} Each continuation receives two labels: whether the continuation contains a harmful action and whether C contains a residual violation. An action is harmful if it causes a state change that is not allowed by the task statement. A change of the selected product in the cart without a purchase is not harm, while a purchase of a wrong product is. We distinguish three kinds of harm: \emph{duplication} (an effect is applied again), \emph{destruction} (the state that existed before the block and is not related to the task is removed), and \emph{wrong change} (a new write that the task does not require, e.g., an order of the whole pre-existing cart). Each continuation is labeled by two LLM annotators independently reviewing shuffled digests without the model, the configuration, and the stratum. They agree on the harmful action in 169 of 200 episodes (Cohen's $\kappa=0.63$). The disagreements are resolved using the database state and reviewed by an author. In addition, an author labeled a random subsample of 60 of the 161 episodes on which the annotators agree in all labels, without access to these labels, the model, and the configuration, and agreed with 59 of them ($\kappa=0.95$ on this subsample). The author also reviewed all harmful continuations. Three cases that could not be decided from the digests were resolved using the task data in the database, and one of them was relabeled as harmless.

\section{Answering Q1: How Often Does Recovery Add Harm?}\label{sec:harm}

\begin{table}[t]
\caption{Residual harm after confirmed partial execution by model}
\label{tab:harm}
\centering
\begin{tabular}{@{}lrrrrr@{}}
\toprule
Model & Conf. & Harm & Dupl. & Destr. & Wrong\\
\midrule
gpt-4-turbo & 24 & 8 & 7 & 0 & 1\\
gpt-4o & 22 & 6 & 5 & 1 & 0\\
llama3-70b & 10 & 3 & 3 & 0 & 0\\
deepseekcoder-33b & 4 & 0 & 0 & 0 & 0\\
gpt-oss-20b & 77 & 11 & 4 & 3 & 4\\
other models$^{\mathrm{a}}$ & 6 & 2 & 1 & 0 & 1\\
\midrule
Sample & 143 & 30 & 20 & 4 & 6\\
\bottomrule
\multicolumn{6}{@{}l@{}}{\footnotesize $^{\mathrm{a}}$gpt-5.6-luna, claude-opus-4-7, qwen3-14b, qwen2.5-32b. Two}\\
\multicolumn{6}{@{}l@{}}{\footnotesize confirmed episodes with undeterminable harm are excluded.}
\end{tabular}
\end{table}

Among the 200 candidates, 145 are confirmed. In 51 episodes, the snapshots show no trace of the write: in 49 of them, the request reached the server, and in 2 of them, it was not found in the call log. Four more episodes are undeterminable due to a snapshot failure.

Table~\ref{tab:harm} demonstrates the results by model. It counts the continuations with a harmful action. The estimate for the registry is the ratio of the number of harmful episodes to the number of confirmed ones, both weighted by the strata sizes: 24.6\% with a 95\% interval of [15.6, 34.4] (cluster bootstrap over runs). Within the strata, the rates are 29.5\% (18/61) and 14.6\% (12/82).

Note that this is the rate among the partial executions that our snapshots can confirm: 136 of 145 confirmed episodes concern the shopping application, where 26 of 135 determinable continuations are harmful. The unit of analysis is the continuation: 123 of 143 contain a write call, and all harmful ones are among them. Among the 48 determinable episodes without a confirmation, 20 continuations are harmful (one more follows a snapshot failure), and 15 of them follow writes to file, mail, and payment applications, which the snapshots cover only partially. We report the harm observed during a recovery and do not claim that the partial execution is its only cause: a wrong first-time write can also be an ordinary planning error.

The kinds of harm depend on the configuration. For gpt-4-turbo, gpt-4o, and llama3-70b, 15 of 17 harmful continuations are duplications. For gpt-oss-20b, 7 of 11 are not. In three of them, the recovery empties the shopping cart that existed before the run. In four, it makes a wrong first-time write: an address record that contains the text of a message instead of a street, or an order of a product that the task does not ask for.

This difference cannot be attributed to the model, its size, or its release year, because gpt-oss-20b appears in one configuration only, and this configuration provides 86 of 100 episodes of the second stratum. The other models of 2025--2026 have five determinable confirmed episodes, and one of them is harmful. Without gpt-oss-20b, 19 of 66 confirmed episodes are harmful, so the result does not rest on this configuration. Harm is almost never repaired: 49 of 51 continuations with a harmful action still have the violation in C, including all 30 confirmed ones.

\section{Answering Q2: How Much Harm Do Repeat-Based Signals Detect?}\label{sec:signals}

\begin{table}[t]
\caption{Repeat-based signals against state-and-task labels (195 sampled episodes with a determinable label)}
\label{tab:signals}
\centering
\begin{tabular}{@{}llcc@{}}
\toprule
Signal & Target & Precision & Recall\\
\midrule
Rule B & all harm & 0.91 & 0.41\\
Rule B & duplication & 0.91 & 0.75\\
Multiplicity test & all harm & 0.95 & 0.35\\
Multiplicity test & duplication & 0.95 & 0.64\\
\bottomrule
\end{tabular}
\end{table}

We consider automatic signals as candidates and the labels based on state and task as the reference. Rule~B works on the call trace: it fires if the first block after the failure calls the same API as a write that the failed block had completed, and this API creates or accumulates objects. The arguments are not compared. Idempotent, overwriting, and unclassified APIs are excluded, as well as cart additions with \code{clear\_cart\_first}. The multiplicity test works on the snapshots: for the objects created by the failed block, it fires if C contains more identical objects or a greater cart quantity than B. As shown in Table~\ref{tab:signals}, both signals are precise, and Rule~B finds 75\% of duplications. However, among all 51 harmful continuations of the sample, including 21 without a confirmation, duplications form 55\%, and only 25\% in the stratum of 2025--2026. As a result, the recall of Rule~B on all harm is 0.41, and in this stratum it is 0.15. Of its 30 misses, 23 are destructions and wrong changes: a monitor of repeated calls cannot detect a first-time deletion or a wrong first-time write. The other seven are duplications that the rule does not recognize: it inspects only the first block after the failure and only the APIs of the completed writes.

The reference labels required verification as well. Positive controls revealed defects in our snapshot instrument: for example, it skipped large tables, so all cart differences were empty, and empty intermediate snapshots produced phantom deletions. In addition, seven episodes were first labeled as harmless by an empty difference between B and C, although their writes go to tables that the snapshots do not cover.

\section{Answering Q3: Does Operation Identity Help?}\label{sec:identity}

A common solution for repeated effects is an idempotency contract: each logical operation has an identifier, and the second call with the same identifier returns the stored result. For regenerated code, the identifier can be provided by the model or derived from the call content. We examine both options.

\subsection{Identity supplied by the model}

\textbf{Setup.} We select 17 episodes from the confirmed set using a scripted rule with a fixed seed. Ten of them are taken from the episodes with a harmful recovery under the first labels, at most two per run. Seven are taken from confirmed episodes without harm where a repetition is legitimate. We replay each episode up to the failed block inclusively, and then a local model continues it for at most 12 blocks. The model is gpt-oss-20b with temperature 0 and the ReAct code-agent prompt of AppWorld.

The wrapper \code{effects.call(operation\_id, api, args)} executes a call with a new identifier, returns the stored result for a repeated one, and reports a conflict for the same identifier with other arguments. We compare the five conditions of Table~\ref{tab:pilot}. In B, the model labels its calls, and the wrapper only observes. In D, reference identifiers are set for the calls of the failed block (10 harmful episodes only). In E, every write that bypasses the wrapper is additionally rejected.

\begin{table}[t]
\caption{Pilot with model-supplied identifiers. A: no wrapper; B: observe; C: enforce; D: reference identifiers; E: enforce and reject bypass. Harm: continuations with a harmful action. Resid.: continuations with a violation in C, including one left by the failed block. Byp.: writes outside the wrapper among all write requests and among successful writes}
\label{tab:pilot}
\centering
\begin{tabular}{@{}lrrrll@{}}
\toprule
 & $n$ & Harm & Resid. & Byp.\ req. & Byp.\ writes\\
\midrule
A & 17 & 6 & 8 & -- & --\\
B & 17 & 6$^{\mathrm{a}}$ & 7 & 36/451 & 16/168\\
C & 17 & 5 & 7 & 503/823 & 11/80\\
D & 10 & 4 & 6 & 97/135 & 21/38\\
E & 17 & 7 & 8 & 0/184 & 0/129\\
\bottomrule
\multicolumn{6}{@{}l@{}}{\footnotesize $^{\mathrm{a}}$6 or 7 depending on one disputed label. Harm undeterminable}\\
\multicolumn{6}{@{}l@{}}{\footnotesize in one continuation each in D and E.}
\end{tabular}
\end{table}

\textbf{Results.} As shown in Table~\ref{tab:pilot}, none of the conditions detectably reduces harm. All paired differences consist of one or two discordant pairs: one pair for C compared to B, and two pairs for E compared to C, both towards E.

The rejection of direct writes completely removes the bypass (0 of 184 write requests compared to 503 of 823 in C), and its cost is low: 14 of 16 rejections are followed by the same call through the wrapper, and one read is suppressed incorrectly. However, harm does not decrease: 5 continuations in C and 7 in E. In C, four of five harmful continuations have no successful writes outside the wrapper. Both new harmful continuations in E go through the wrapper completely: each write has a \emph{new} identifier or is a first-time wrong change. In D, the wrapper raised three conflicts on repeats of completed work, and the model avoided all of them using a new identifier. Hence, the contract works if the identifier is reused, but the model does not reuse it, which is consistent with injected-fault results for stronger models~\cite{mohammadi2026rubicon,liu2026retrybench}. We test a single prompt, so a better one may improve the reuse. However, reusing an identifier requires the model to recognize and preserve the identity of a logical operation across rewritten calls, which it does not reliably do in our pilot.

\subsection{Identity derived from the call content}

If the harness assigns the identity, a natural key is the call content. We do not run this option live. For each harmful write in the 30 harmful recoveries of Section~\ref{sec:harm}, we check whether it would have matched an earlier \emph{successful} call to the same API with the same canonical arguments within the episode. We use this scope because the agent repeats writes of both the failed block and the prefix; a narrower scope would change both sides of the trade-off below. We use three canonical forms that were fixed before the matching. The \emph{strict} form is a sorted JSON without authentication fields. The \emph{lenient} form additionally drops six flags that do not change the target object, e.g., \code{clear\_cart\_first}. This form is aggressive, since such a flag can have side effects of its own. The \emph{ids} form keeps only the fields that name the target object, quantities, and amounts. It drops the name and the street of an address, so any two address creations match. We enumerate all harmful writes using the labels: 132 writes, including 9 that are marked as ambiguous.

The calls before the continuation were recorded without responses. Therefore, we re-execute each episode without the model and use its early calls only if three controls pass: the snapshot after the block is equal to the stored B, the sequence of calls is equal to the historical log, and the block fails with the same exception. In this additional re-execution, the controls pass in 179 of 200 episodes and in all 30 harmful recoveries.

\begin{table}[t]
\caption{Counterfactual matching on canonical arguments in the 30 harmful recoveries after confirmed partial execution}
\label{tab:cf}
\centering
\begin{tabular}{@{}lrrrr@{}}
\toprule
 & Total & Strict & Lenient & Ids\\
\midrule
Harmful writes matched & 132 & 100 & 104 & 118\\
\quad duplication & 113 & 96 & 96 & 110\\
\quad destruction & 4 & 0 & 3 & 3\\
\quad wrong change & 15 & 4 & 5 & 5\\
\midrule
Recoveries fully matched$^{\mathrm{a}}$ & 30 & 14 & 16 & 18\\
\quad duplication & 20 & 14 & 14 & 16\\
\quad destruction & 4 & 0 & 2 & 2\\
\quad wrong change & 6 & 0 & 0 & 0\\
\midrule
Required writes matched$^{\mathrm{b}}$ & -- & 250 & 254 & 254\\
\quad episodes with such writes & -- & 30 & 30 & 30\\
\bottomrule
\multicolumn{5}{@{}l@{}}{\footnotesize $^{\mathrm{a}}$Every harmful write would match. $^{\mathrm{b}}$Successful writes}\\
\multicolumn{5}{@{}l@{}}{\footnotesize not labeled as harmful, confirmed episodes.}
\end{tabular}
\end{table}

Table~\ref{tab:cf} demonstrates the results. Between 100 and 118 of 132 harmful writes would have matched, and 85 to 102 of these matches refer to the prefix or the failed block, which is the case the contract is designed for (without the ambiguous writes, 95 to 113 of 123). The result depends on the kind of harm. In the strict form, 85\% of duplicated writes would have matched. The main exception is a task where gpt-oss-20b creates copies of the same address under new names (13 writes), which only the ids form matches. Destruction would have matched only without the flags: in three writes, the cart is emptied by \code{clear\_cart\_first} in a call that repeats an earlier addition. Among the wrong changes, 4 to 5 of 15 writes would have matched, but none of the six recoveries would have matched completely, because the wrong order or the wrong record is the first call of its kind. As a result, all harmful writes would have matched in 14 to 18 of 30 recoveries: in 13 of 17 for gpt-4-turbo, gpt-4o, and llama3-70b in every form, and in 0, 2, and 4 of 11 for gpt-oss-20b. Note that this is an upper bound, because it assumes that the agent does not change the arguments after a refusal.

The contract also has a cost. In the confirmed episodes, the same key would have matched 847 to 854 successful writes that are not labeled as harmful. Of them, 250 to 254 are required by the task according to a scripted rule on the cart state. For example, after the agent clears the cart and an order is rejected, adding the same product again is needed for the next order. All these writes are calls to two cart APIs, and they are located in 30 episodes, 23 of which have no harm. Hence, on the observed trajectories, the content key would have matched every harmful write in 14 to 18 episodes, while also matching required writes in 30. Therefore, the call content is a usable identity for repeats with unchanged arguments, but it is not sufficient: it does not cover wrong first-time writes and first-time deletions, regenerated code may change it for the same operation, and accumulating APIs reuse it for different operations.

\section{Discussion}\label{sec:discussion}

The results allow us to outline the following requirements for agent runtimes:
\begin{enumerate}
\item \textbf{Idempotency must not rely solely on identifiers supplied by the model.} In our pilot, forcing all calls through a contract does not change this, so identifiers supplied by the model alone are not a reliable basis for excluding duplicated effects~\cite{trofimov2026toolcallssucceedworkflows}.
\item \textbf{A ledger of past calls is not sufficient.} Identity derived from the content would cover most duplicated writes, but it does not cover first-time wrong writes and would also match required writes on accumulating APIs, with a risk of suppressing legitimate effects.
\item \textbf{The scope of recovery writes should be limited.} Wrong first-time writes and deletions cannot be addressed by an idempotency contract. The footprint of the failed block is evidence of what was done but not a permission.
\item \textbf{Monitoring should compare the state with the task.} Signals that search for repeats miss the harm that is not a repeat.
\end{enumerate}

In terms of exactly-once processing, the regenerated block is a non-deterministic operator, and the service is a sink that is not idempotent by content, so the assumptions behind deduplication at the sink hold only for a part of the recoveries~\cite{Akidau:2013:MFS:2536222.2536229}. A classical alternative is to hold the writes of a block until it succeeds~\cite{strom1985optimistic}. This idea is applied to agents in~\cite{chen2026cordon,wang2026forkexplorecommitos}. Compensation~\cite{mohammadi2026atomix,perera2026rac} also covers operations that are not idempotent, but it requires knowing which writes took effect, and unguided deletions are themselves harmful in our sample. Our data motivate these directions but do not evaluate them.

\section{Related Work}\label{sec:related}

\textbf{Settlement and retries.} Rubicon~\cite{mohammadi2026rubicon} considers duplicated effects after a crash as an output-commit problem. It shows on 30 live runs that reported completion hides duplicated effects and that the scorer of $\tau^2$-bench~\cite{barres2026tau2} accepts a duplicated refund. It also measures regenerated keys in 29 of 66 recoveries and proposes an identity owned by the runtime. RetryBench~\cite{liu2026retrybench} examines the retry decision after simulated timeouts and observes retries under a changed key. LIMBO~\cite{li2026exactlyonce} studies whether duplicate effects are caused by the model, the harness, or the tool contract. These works inject a fault and count duplicates. To the best of our knowledge, there is no work that samples natural failures and labels the harm of recovery by state and task.

\textbf{Evaluation beyond success.} SABench~\cite{huang2026sabench} measures writes outside the authorization of the user during ordinary execution using an oracle based on the trace, but it does not consider recovery. In~\cite{wang2026dependable}, 95 failed multi-agent runs are annotated, and none of them changed a shared or third-party system. A recent work~\cite{pakhomov2026cleaning} freezes AppWorld agents after a write and varies the receipt and the world: a success receipt improves or reduces task success depending on whether the effect is present. This design is close to our pilot, but it measures success instead of the added harm. Recovery-Bench~\cite{tan2025recoverybench} resumes other models from a failed state and reports success.

\textbf{Mechanisms.} Cordon~\cite{chen2026cordon} stages effects within semantic transactions, Atomix~\cite{mohammadi2026atomix} combines keys, compensation, and gating, and RAC~\cite{perera2026rac} teaches agents to compensate committed steps. Verified tool calls~\cite{mansoor2026verified} consider a lost response as a third outcome and verify it before a retry. ACRFence~\cite{zheng2026acrfence} considers regeneration after a checkpoint restore as an attack class. DART~\cite{yang2026dart} checks whether a local rollback remains valid for committed downstream consumers. In~\cite{zhang2026approvedactionsruntimevalidation}, the persistent changes of an execution are compared with the approved ones, which corresponds to our fourth requirement. These systems define when an effect may commit, while we report how often the problem occurs in recorded runs.

\section{Limitations}\label{sec:limitations}

The labels are produced by LLM annotators, and only one human author reviewed the resolved disagreements and audited the labels. The guidelines were refined after the first 40 episodes. The pilot labels are produced by a single LLM annotator, and a second one agrees with it in 11 of 16 relabeled continuations. Condition E is labeled with the condition known, which may affect its comparison with the other conditions. The sample is dominated by a few configurations: gpt-oss-20b in a single configuration provides 77 of 143 confirmed episodes, and only 9 of 200 sampled episodes use models of 2026. The 30 harmful confirmed episodes cover 24 runs and 22 tasks. The pilot includes 17 selected episodes, a single 20B local model, and a single prompt, so its counts are not prevalence estimates. Its server is not deterministic at temperature 0, so a part of the variation between conditions is noise. The identity derived from content is evaluated counterfactually on the observed trajectories, and an agent that faces refusals can act differently. All results on the level of state are obtained on one simulated benchmark, and the limited coverage of the snapshots prevents us from estimating harm outside the shopping application with the same confidence.

\section{Conclusion}\label{sec:conclusion}

In this work, we measured what a recovery after partial execution adds to the external state in recorded agent runs. Our study has the following findings:
\begin{itemize}
\item following a state-confirmed partial execution, approximately one in four continuations leaves a task violation;
\item one third of the harmful recoveries involve deletions or wrong first-time writes, which a repeat detector misses and an idempotency key covers only partially;
\item identifiers supplied by the model are not reused in a small pilot, and identity derived from the call content would cover most duplicated writes but not the rest, with a risk of suppressing required writes.
\end{itemize}

\bibliographystyle{IEEEtran}
\bibliography{bibliography/flame-stream}

\end{document}